\pdfoutput=1
\documentclass[11pt]{article}

\usepackage[utf8]{inputenc}
\DeclareUnicodeCharacter{1E24}{\d{H}}
\DeclareUnicodeCharacter{201B}{`}

\usepackage[final]{acl}

\usepackage{times}
\usepackage{float}
\usepackage{latexsym}
\usepackage{booktabs}
\usepackage{amsmath}
\usepackage{hyperref}    % for clickable URLs
\usepackage{amssymb}     % for \checkmark
\usepackage{tipa}
\usepackage{multirow}
\usepackage{xcolor}
\definecolor{mypink}{rgb}{0.9686274,0.882352,0.86666}
\definecolor{mygreen}{rgb}{0.819607,0.890196,0.760784}
\usepackage{times}
\usepackage{latexsym}
\usepackage{graphicx}
\usepackage{float}
\usepackage{colortbl}
\usepackage{multirow}
\usepackage{booktabs}
\usepackage{subcaption}
\usepackage{rotating}
\usepackage{placeins}
\usepackage{comment}
\usepackage{pifont}

\newcommand{\cmark}{\ding{51}} % check mark
\usepackage[export]{adjustbox}
\usepackage{dirtytalk} %correct quotation marks in any language

\usepackage{algorithm}
\usepackage{algpseudocode}
\usepackage{float}
\usepackage{pdflscape}
\usepackage{hyperref}
\usepackage[]{acronym}
\usepackage{amsmath}

\usepackage{amssymb}
\usepackage[T1]{fontenc}
\usepackage[utf8]{inputenc}

\usepackage{microtype}

\usepackage{inconsolata}

\usepackage{graphicx}

\title{Modern Models, Medieval Texts: A POS Tagging Study of Old Occitan} % First idea, please feel free to change!

\author{
  \textbf{Matthias Schöffel\textsuperscript{1,2}},
  \textbf{Marinus Wiedner\textsuperscript{3}},
  \textbf{Esteban Garces Arias\textsuperscript{2,4}},
    \textbf{Paula Ruppert\textsuperscript{2},}\\
  \textbf{Christian Heumann\textsuperscript{2}},
  \textbf{Matthias Aßenmacher\textsuperscript{2,4}}
\\
\\
    \textsuperscript{1}Bavarian Academy of Sciences,
  \textsuperscript{2}LMU Munich,
  \textsuperscript{3}Albert-Ludwigs-Universität Freiburg,\\
  \textsuperscript{4}Munich Center for Machine Learning (MCML)\\
\\
  \small{
    \textbf{Correspondence:} \href{mailto:esteban.garcesarias@stat.uni-muenchen.de}{matthias.schoeffel@badw.de}
  }
}

\begin{document}
\maketitle
\begin{abstract}
Large language models (LLMs) have demonstrated remarkable capabilities in natural language processing, yet their effectiveness in handling historical languages remains largely unexplored. This study examines the performance of open-source LLMs in part-of-speech (POS) tagging for Old Occitan, a historical language characterized by non-standardized orthography and significant diachronic variation. Through comparative analysis of two distinct corpora—hagiographical and medical texts—we evaluate how current models handle the inherent challenges of processing a low-resource historical language. Our findings demonstrate critical limitations in LLM performance when confronted with extreme orthographic and syntactic variability. We provide detailed error analysis and specific recommendations for improving model performance in historical language processing. This research advances our understanding of LLM capabilities in challenging linguistic contexts while offering practical insights for both computational linguistics and historical language studies.
\end{abstract}

\begin{figure}[t!]
\centering
\subfloat[Map of the Occitan-speaking region in southern France, northeastern Spain, and northwestern Italy.]{%
  \includegraphics[width=79mm,keepaspectratio]{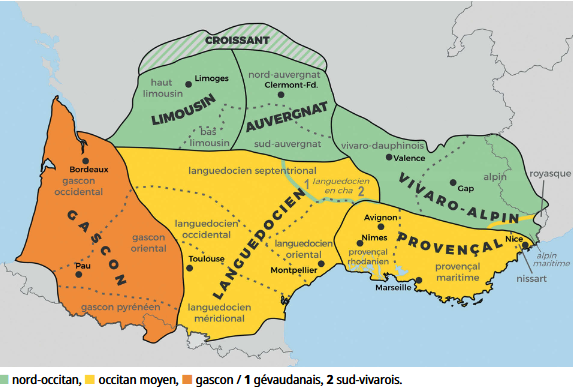}
  \label{fig:figure_one_a}
}

\vspace{1em} % vertical space between subfigures

\subfloat[Graphical variations in spelling, exemplified by the term \textit{abeurador}, highlighting the challenges posed by non-standardized orthography.]{%
  \includegraphics[width=80mm,keepaspectratio]{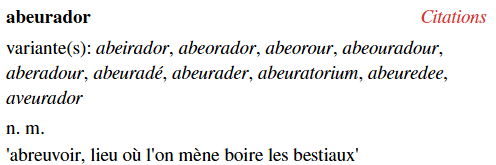}
  \label{fig:figure_one_b}
}

\caption{(a) Geographic distribution of Old Occitan with its principal dialect zones \citep{sibille:2024}. (b) Orthographic diversity in Old Occitan texts, as evidenced by multiple graphical variants of the same term, illustrating inherent challenges for modern LLMs.}
\label{fig:figure_one}
\end{figure}

\section{Introduction}
\label{intro}

Old Occitan, also known as Old Provençal, was widely spoken from the 11th to the 16th century across southern France, northeastern Spain, and northwestern Italy (cf. Fig. \ref{fig:figure_one}(a)). This language played a pivotal role in shaping both Romance linguistics and medieval European literature, particularly through its renowned troubadour tradition. However, computational analysis and digital preservation of Old Occitan face significant challenges, primarily due to the limited availability of digitized manuscripts and annotated corpora compared to contemporary medieval languages such as Old French \citep{konvens:58_scrivner12w}. A key obstacle in processing Old Occitan texts is their pronounced orthographic variation, as illustrated in Figure \ref{fig:figure_one}(b) through the term \textit{abeurador} ('watering place'), which exhibits substantial regional and textual variations in spelling. These variations, while historically significant, present particular challenges for automated text processing tasks such as Part-of-Speech (POS) tagging, which is the focus of the present work.

The imperative for accurate POS tagging in low-resource languages like Old Occitan extends beyond mere technical curiosity. POS tagging is a foundational step in numerous NLP applications, from syntactic parsing and information extraction to more advanced tasks in digital humanities. For historical languages, reliable tagging is critical not only for linguistic analysis but also for reconstructing the evolution of language, understanding regional variation, and supporting interdisciplinary research that bridges history and computational methods. Moreover, the performance of LLMs on such texts offers insights into the adaptability of modern models when confronted with non-standardized data -- a challenge that remains largely unaddressed in contemporary NLP research.

In this study, we systematically evaluate a range of LLMs using various prompting strategies—Zero-shot, few-shot, and instruction-tuned approaches—on a corpus comprising 91,953 tokens. Beyond a mere exploration of current capabilities, our work elucidates key factors influencing model performance and offers a rigorous error analysis and practical recommendations to mitigate the effects of input modifications and enhance POS tagging accuracy.

\noindent \textbf{Research Questions:} Our study addresses the following research questions: (i) \textbf{RQ1:} How effectively can current LLMs perform POS tagging on Old Occitan texts, given the challenges posed by non-standardized orthography and sparse annotated resources? (discussed in Section \ref{llm_perf}) (ii) \textbf{RQ2:} Which prompting strategies—Zero-shot, few-shot, or instruction-tuned—yield the most robust performance on this low-resource, historical language? (discussed in Section \ref{prompt_perf}) (iii) \textbf{RQ3:} What specific error patterns and model biases emerge during POS tagging, and how can these insights inform practical improvements? (addressed in Sections \ref{sec:erroranalysis} and \ref{sec:recommendations}). By answering these questions, we aim to bridge modern NLP techniques with the nuanced demands of historical linguistics.

\noindent \textbf{Contributions:} We summarize our contributions as follows:
\begin{enumerate}
    \item We provide the first comprehensive evaluation of multiple LLMs for POS tagging on Old Occitan texts, establishing a robust baseline for historical Romance languages.
    \item We systematically compare concrete prompting strategies—including Zero-shot, few-shot, and instruction-tuned approaches—to adapt LLMs to the irregularities of non-standardized historical data.
    \item We perform a detailed error analysis to uncover model-specific biases and limitations, offering targeted recommendations to improve POS tagging performance on low-resource texts.
    \item We release a novel POS Tagging dataset for Old Occitan, along with our code and experimental results, to facilitate future research in historical NLP.\footnote{\href{https://github.com/msch38/occ_pos_tagging}{Link to our GitHub Repository.}}
\end{enumerate}

\section{Related work}
\label{related}

Part-of-speech (POS) tagging for low-resource languages presents unique challenges that have gained increasing attention in computational linguistics. Several approaches have emerged to address data scarcity in these settings, with varying degrees of success.
\citet{cardenas-etal-2019-grounded} proposed a grounded unsupervised universal POS tagger for low-resource languages, framing tagging as a clustering problem followed by decipherment-based grounding. This approach requires no labeled training data and demonstrates reasonable performance across diverse languages. Building on this work, \citet{plank2018bestworldslexicalresources} demonstrated that integrating conventional lexical information can significantly improve neural cross-lingual POS tagging, suggesting that even small amounts of symbolic lexical resources can be valuable when gold-standard corpora are unavailable.
However, \citet{kann2020weakly} challenged the effectiveness of weakly supervised approaches for truly low-resource languages. Their evaluation across 15 typologically diverse languages revealed that state-of-the-art weakly supervised POS taggers perform significantly worse under realistic resource constraints than previously reported, with accuracy below 50\% for most languages. This skepticism is further supported by \citet{moeller-etal-2021-pos}, who found that the presence or absence of POS tags does not significantly impact performance in morphological learning tasks, with some cases showing improved performance when POS tags were removed.
For endangered languages specifically, \citet{Anastasopoulos2018PartofSpeechTO} evaluated POS tagging techniques on Griko, achieving 72.9\% accuracy through combined semi-supervised methods and cross-lingual transfer. Similarly, \citet{10074031} demonstrated success with the endangered Indian tribal language Katkari, achieving 86.84\% accuracy using Hidden Markov Models and the Viterbi algorithm, suggesting that traditional statistical approaches remain viable for low-resource scenarios.
Recent work has focused particularly on languages with dialectal variation. The creation of CorpusArièja by \citet{poujade-etal-2024-corpusarieja} provides a valuable resource for Occitan, containing 41,000 tokens with POS tags and handling both dialectal and spelling variations. Building on this, \citet{hopton-aepli-2024-modeling} demonstrated that large multilingual models can effectively handle dialectal variation in Occitan without requiring spelling normalization, particularly when fine-tuned for POS tagging. More recently, there have been efforts to ramp up the availability of resources for Old Occitan, including the creation of a digital version of the Old Occitan dictionary \footnote{\textit{DOM: Dictionnaire de l'occitan médiéval} \\ \url{http://www.dom-en-ligne.de/}}at the Bavarian Academy of Sciences. Building on handwritten resources, \citet{arias-etal-2023-automatic} tackled automatic transcription, combining a custom-trained Swin image encoder with a BERT-based text decoder to enhance digitization of Old Occitan spelling variations.

\section{Data}
\label{data}

%\textbf{TODO:} Describe datasets (age, size, context) as well as the data splitting protocol.

Our benchmark comprises two corpora drawn from distinct domains: a hagiographical text and a medical treatise. The former is represented by the \textit{Vida de Sant Honorat}, while the latter is embodied by \textit{On surgery and instruments} by Abū l-Qāsim al-Ḥalaf al-Zahrāwī (Albucasis).

For the hagiographical corpus, the primary source is the manuscript Nouvelle Acquisition Fran\c{c}aise 6195 (NAF6195, also known as manuscript M of the \textit{Vida de Sant Honorat}), preserved at the Bibliothèque Nationale de France. Dated to the 14th century and originating from Provence, this manuscript was first digitised following an archival visit. Its contents were then semi-automatically transcribed using a handwritten text recognition model specifically developed for Old Occitan scripts \cite{Wiedner2023} and subsequently subjected to rigorous manual revision. A pre-annotation step was performed with a modern Occitan part-of-speech tagger \cite{PoujadeInProgress}, after which manual corrections were again applied. The final corpus comprises 44,044 tokens and, to our knowledge, has not previously underpinned any extant editions of the \textit{Vida de Sant Honorat}. A notable linguistic feature of this text is the presence of graphical variants that markedly diverge from those catalogued in the DOM (79,840 entries, 38,861 unique lemmas, and 40,979 graphical variants as of February 2025), as detailed in Table \ref{tab:NAF6195_variants}.

In contrast, the medical corpus is derived from \textit{On surgery and instruments} by Albucasis. Originally composed in Arabic as one volume of the thirty-volume medical encyclopedia commonly known as al-Tasrif and dating from the late 10th century, the text encompasses nearly 57 chapters and 42,099 word tokens. It was later translated into Latin by Gerard of Cremona at the Toledo School of Translators (circa 1180 AD) and subsequently into vernaculars, including Old French (mid-13th century) and Old Occitan (second quarter of the 14th century). For our purposes, we employed an existing electronic version of the Old Occitan edition \cite{alb_occ}, originally compiled by P.T. Ricketts, converted to TEI format by Dominique Billy, and released in 2015 under a Creative Commons licence (CC BY-NC-SA 4.0). This edition is based on the manuscript preserved in the Bibliothèque de l'Université (Montpellier), Faculté de médecine, 95. The treatise is distinguished by its specialised technical vocabulary spanning surgery, anatomy, pharmacy, botany, and zoology, and it integrates a mosaic of linguistic influences, including Arabic, Latin, Greek, and vernacular elements. For instance, the Arabic term \textit{taxmir} (connoting ‘blepharoplasty’)—derived from \textit{tašmir}—is attested in several graphical variants (e.g. atactini, ataxmir, tactimi, tactinir, taxanir).

\begin{figure}[ht]
\centering
\includegraphics[width=80mm,keepaspectratio]{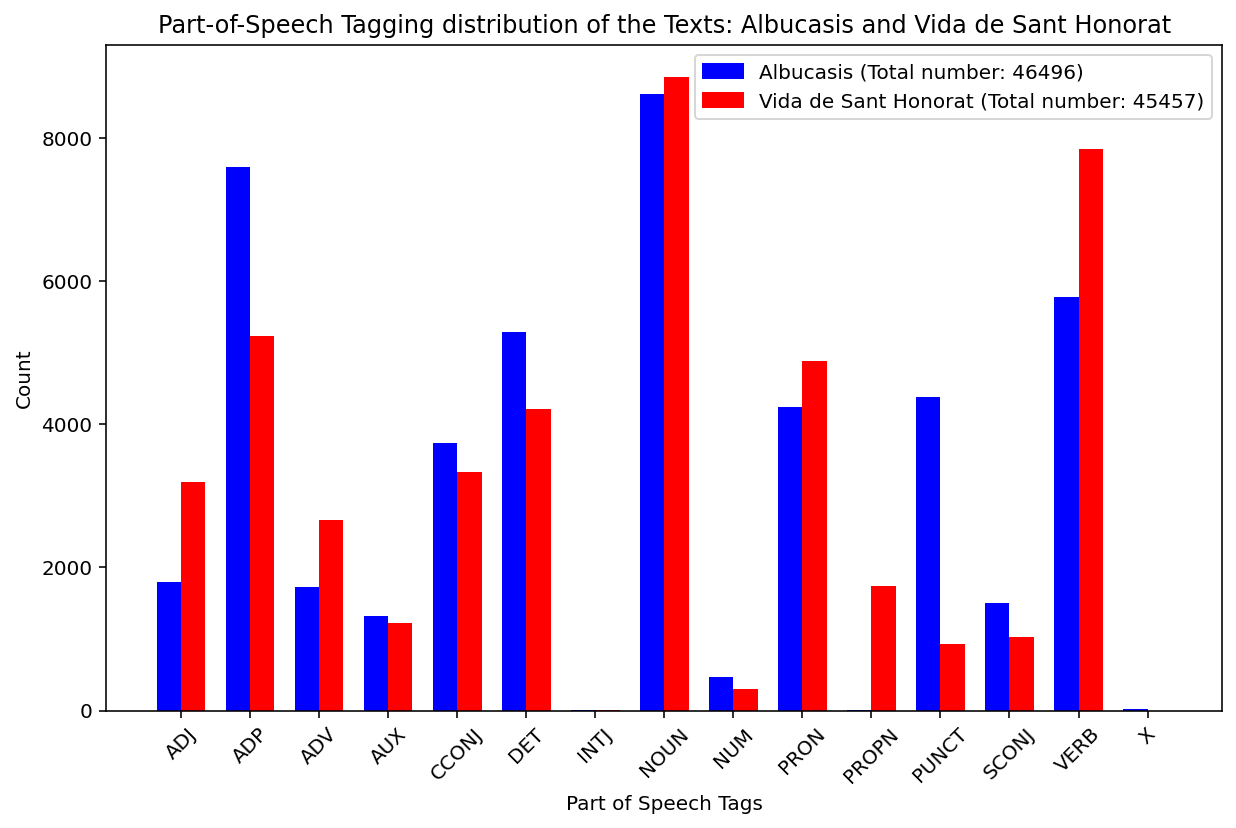}
\caption{Part-of-Speech distribution for both texts: Albucasis (blue) and Vida de Sant Honorat (red).}
\label{fig:figure_two}
\end{figure}

Both texts were manually annotated following the Universal Dependencies framework\footnote{\url{https://universaldependencies.org/u/pos/}}. The annotation scheme was constrained to 15 part-of-speech categories (ADJ, ADP, ADV, AUX, CCONJ, DET, INTJ, NOUN, NUM, PRON, PROPN, PUNCT, SCONJ, VERB, and X) owing to the absence of the PART and SYM classes in both corpora. Figure \ref{fig:figure_two} illustrates the part-of-speech distributions across the two texts.

\begin{table}[!ht]
    \centering
    \resizebox{.5\textwidth}{!}{
        \begin{tabular}{ll}
        \hline
        \textbf{New NAF6195 entry} & \textbf{Available DOM entries}  \\ 
        \midrule
        homps (engl. ‘man’) & ome, om, omen, omne, hom, home\\ 
        primpce (engl. ‘prince’) & prince, princep, princip, princer\\ 
        penedensia (engl. ‘penitence’) & penedensa, pendensa, pentensa\\
        ompnipotent (engl. ‘allmighty’) & omnipotent, omnipoten\\
        \bottomrule
    \end{tabular}
    }
    \caption{Graphical variants vs. known (DOM) entries.}
    \label{tab:NAF6195_variants}
\end{table}

\begin{table*}[!ht]
\centering
\resizebox{0.8\textwidth}{!}{
    \begin{tabular}{lccccccccc}
\hline
\textbf{Model} & \textbf{Old Occitan} & \textbf{Occitan} & \textbf{French} & \textbf{Spanish} & \textbf{Italian} & \textbf{Portuguese} & \textbf{Romanian} & \textbf{Arabic} & \textbf{English} \\
\hline
COLaF         &  & \cmark &       &       &       &       &       &       &  \\
Phi4-14B          &        & \cmark & \cmark & \cmark & \cmark & \cmark & \cmark &  \cmark  & \cmark \\
Mistral-7B       &        &        &       &       &       &       &       &       & \cmark \\
Mistral-Nemo-12B  &        &        & \cmark & \cmark & \cmark & \cmark &       &       & \cmark \\
Gemma2-9B     &        &        &       &       &       &       &       &       & \cmark \\
Mixtral-8x7B  &        &        & \cmark & \cmark & \cmark &  &       &       & \cmark \\
Aya-8B        &        &  & \cmark & \cmark & \cmark & \cmark & \cmark & \cmark & \cmark       \\
Qwen2.5-14B   &        &  & \cmark & \cmark & \cmark & \cmark &      & \cmark & \cmark \\
\hline
\end{tabular}
    }
\caption{Overview of supported languages by each model.}
\label{tab:supported_languages}
\end{table*}

\begin{figure*}[!ht]
    \centering
    \includegraphics[width=0.7\textwidth]{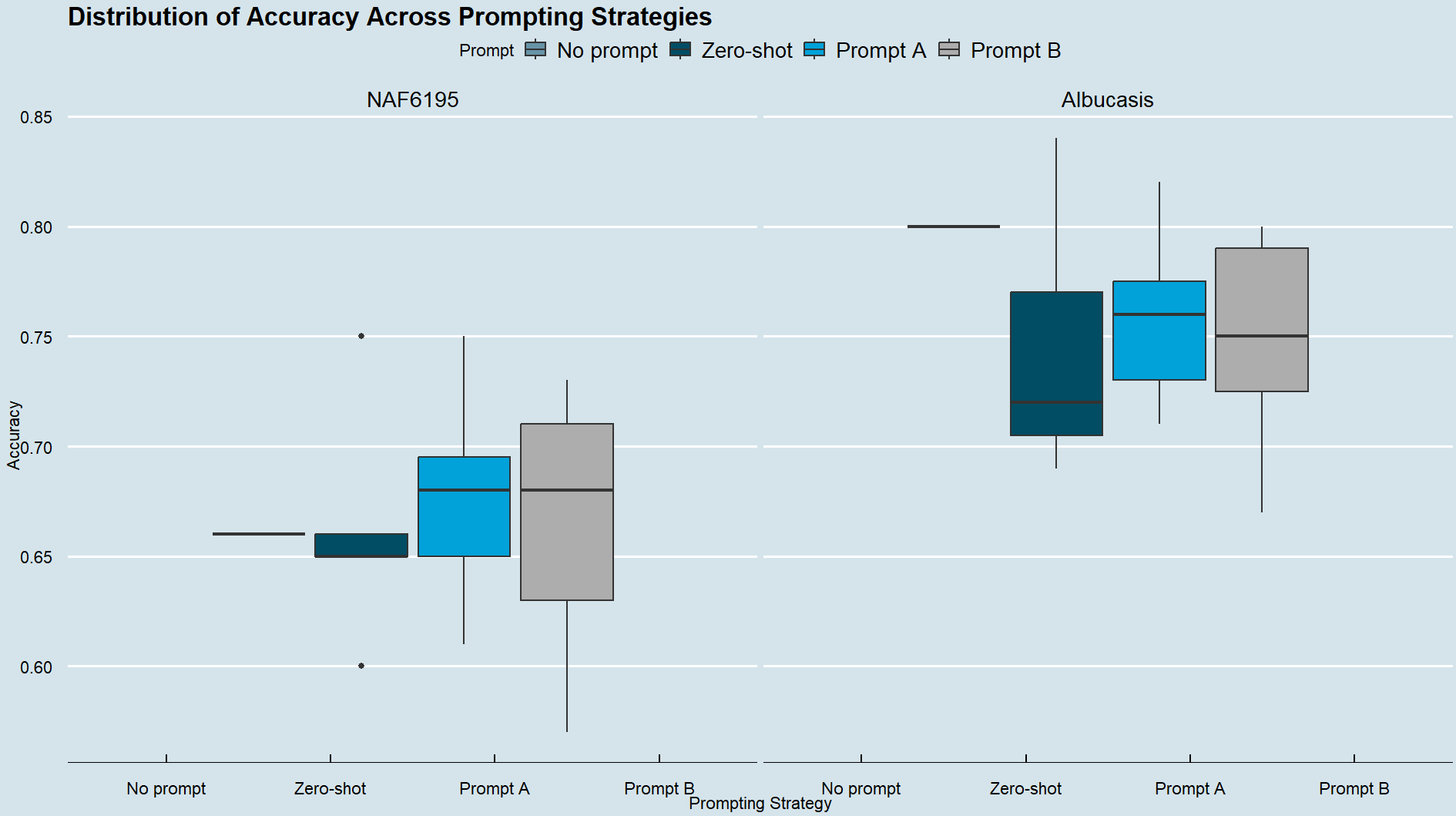}
\caption{Accuracy distribution across different prompting strategies and datasets.}
 \label{fig:prompting_strategies} 
\end{figure*}

\section{Experimental setup}
\label{exp}

\subsection{Models and Hardware}

In this study, we evaluated eight distinct models. Our set comprises the COLaF model \citep{thibault_clerice_2020_3883590, manjavacas-etal-2019-improving, nedey_2024, miletic:hal-02123743}-- a dedicated part-of-speech tagger trained on modern Occitan-- alongside seven open‐source instruct models that exhibit varying levels of support for Romance languages (Fig. \ref{tab:supported_languages}). Specifically, the instruct models include Phi4-14B \citep{abdin2024phi4technicalreport}, Mistral-7B-Instruct-v0.2, Mistral-Nemo-12B, Mixtral-8x7B \citep{jiang2023mistral}, Gemma2-9B \citep{gemmateam2024gemma2improvingopen}, Aya-8B \citep{aryabumi2024aya23openweight}, and Qwen2.5-14B \citep{qwen2025qwen25technicalreport}. Our experiments were conducted employing an NVIDIA Tesla V100-16 GB.

\begin{comment}

COLaF: \url{https://colaf.huma-num.fr/deucalion/occ-cont}
Ollama:
Phi4 (tokenization: tiktoken tokenizer ), mistral, mistral-nemo (Tekken), gemma2-9b (SentencePiece tokenizer with split digits, preserved whitespace, and byte-level encodings) , mixtral8x7b (Byte-fallback BPE tokenizer - ensures that characters are never mapped to out of vocabulary tokens), aya8b, qwen2.5-7b.
\end{comment}

\subsection{Prompting strategies}

We explore three prompting strategies, each increasing in contextual detail and specificity. The simplest approach, \textit{Zero-shot}, directly instructs the model to assign Universal Dependencies Part-of-Speech tags to each word—without any additional context or expert framing. In \textit{Prompt A}, the instructions are enhanced by explicitly positioning the model as a Medieval Occitan language expert. This prompt emphasizes strict token-by-token processing, ensuring that punctuation is preserved and that the order of words remains unchanged. Finally, \textit{Prompt B} builds upon the previous strategies by incorporating rich linguistic context. It provides explicit examples of spelling variations characteristic of Medieval Occitan (such as variations in the spelling of common words), guiding the model to account for these variations during analysis. Table \ref{tab:prompting_strategies} provides a detailed description.

\subsection{Metrics}

To evaluate the performance of LLMs in POS tagging for Old Occitan, we focus on widely-used metrics: Accuracy, Precision, Recall and F1-score. Further, we measure the ratio of correctly POS-tagged phrases. A detailed overview is provided in Appendix, Section \ref{a:metrics}.

\begin{comment}
\subsection{Decoding Strategies}

Once the most promising models and prompting strategies have been selected, evaluate robustness of results and effect of different decoding strategies (e.g. beam search \cite{Freitag_2017}, CS \cite{su2022contrastive}, ACS \cite{garces-arias-etal-2024-adaptive}, sampling with temperature \cite{ackley1985learning}, top-$k$ sampling \cite{fan-etal-2018-hierarchical}, top-$p$ (nucleus) sampling \cite{holtzman2019curious}) and comment about their effect over performance.

\begin{table}[!ht]
    \centering
    \resizebox{.5\textwidth}{!}{
    \input{pos_tagging_decoding_performance}
    }
    \caption{Effect of decoding strategies over POS-tagging performance. Hyperparameters are chosen based on \cite{garces-arias-etal-2025-decoding}.}
    \label{tab:decod_eval}
\end{table}

\end{comment}

\begin{figure*}[!ht]
    \centering
    \includegraphics[width=0.7\textwidth]{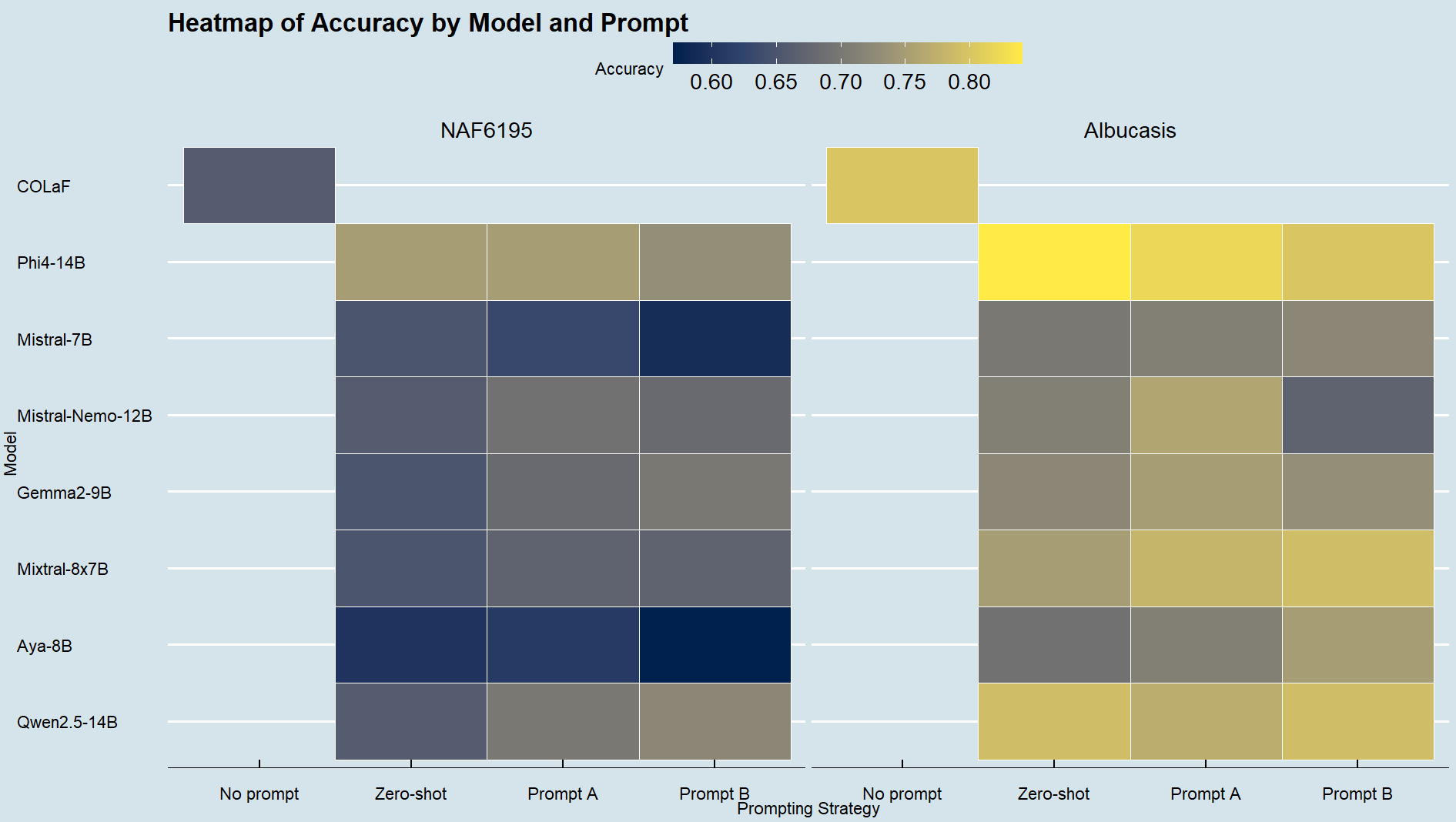}
\caption{Accuracy heatmap for models and prompting strategies. Results on the left correspond to the  NAF6195 dataset and on the right to \textit{Albucasis}.}
 \label{fig:accuracy_heatmap} 
\end{figure*}

\section{Results}
\label{sec:results}

Our extensive evaluation of POS tagging in Old Occitan was performed using two datasets with distinct characteristics. The NAF6195 dataset is annotated from a challenging, non-standardized script with 28\% unknown vocabulary, whereas Albucasis, a publicly available resource, exhibits a slightly lower rate of unknown tokens (25\%). Tables~\ref{tab:pos_tagging_performance_naf6195} and~\ref{tab:pos_tagging_performance_albucasis} provide a comprehensive summary of POS tagging performance for a diverse set of models and prompting strategies.

\subsection{Comparative Performance Across Datasets}
\label{llm_perf}
Overall, the models achieve higher absolute performance on Albucasis compared to NAF6195. For example, the COLaF baseline, which does not utilize prompting, registers an accuracy of 0.80 on Albucasis compared to 0.66 on NAF6195. Similar trends are observed across micro-averaged Precision, Recall, and F1-score. This divergence is likely attributable to the increased orthographic variability and a larger proportion of unknown vocabulary in NAF6195. Figure~\ref{fig:accuracy_heatmap} further highlights this discrepancy by visualizing the distribution of accuracy scores, revealing a broader spread and lower central tendency for NAF6195.

\subsection{Influence of Prompting Strategies}
\label{prompt_perf}
\begin{figure*}[!ht]
    \centering
    \begin{subfigure}[b]{0.48\textwidth}
         \centering
         \includegraphics[width=\textwidth]{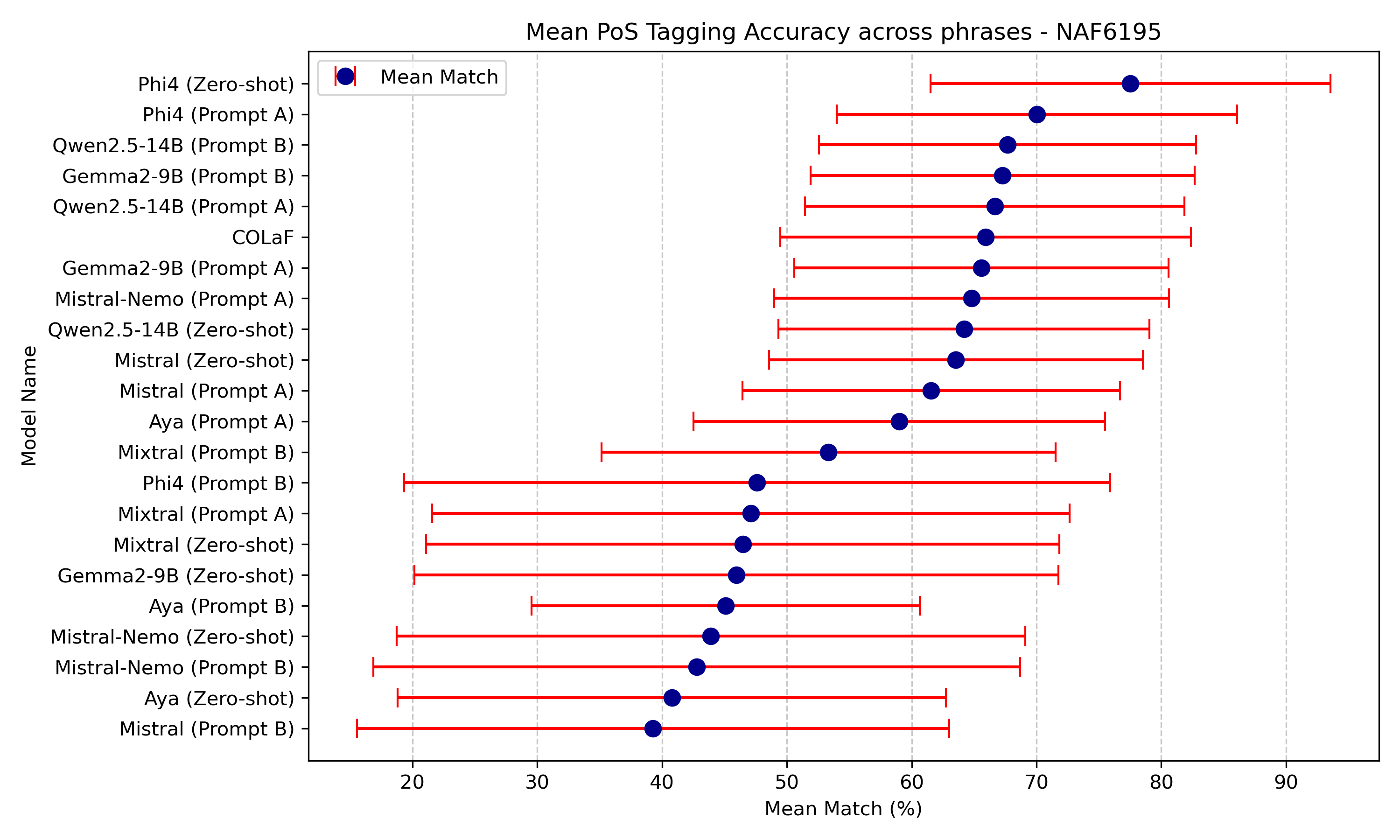}
         \caption{Accuracy vs. Prompt Strategies for the NAF6195 dataset.}
         \label{fig:rcptp_naf}
    \end{subfigure}
    \hfill
    \begin{subfigure}[b]{0.48\textwidth}
         \centering
         \includegraphics[width=\textwidth]{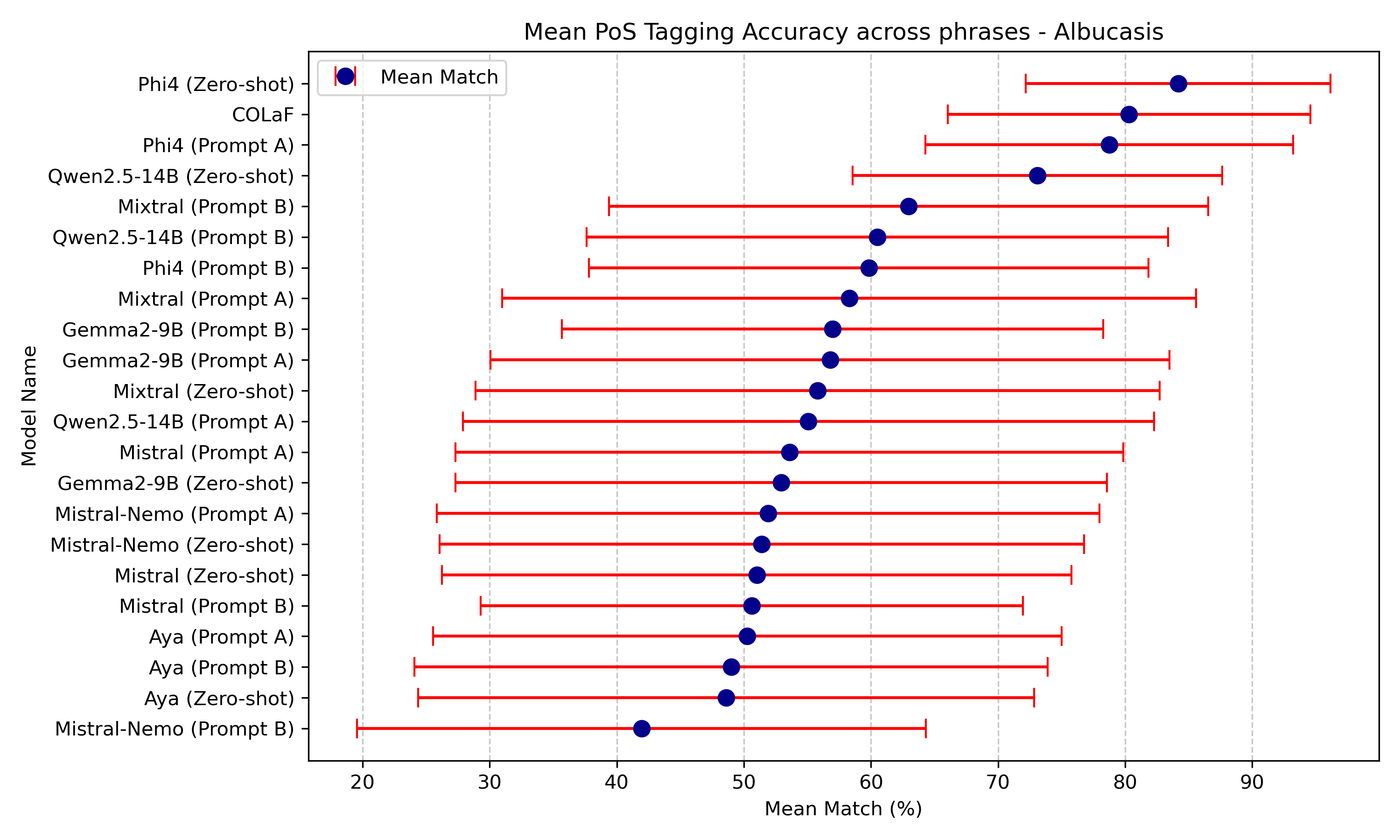}
         \caption{Accuracy vs. Prompt Strategies for the Albucasis dataset.}
         \label{fig:rcptp_albuc}
    \end{subfigure}
    \caption{Accuracy across phrases vs. Prompt Strategies for the NAF6195 and Albucasis datasets.}
    \label{fig:combined}
\end{figure*}

Three prompting configurations were examined: Zero-shot, Prompt A, and Prompt B. In the NAF6195 dataset, a progressive increase in median accuracy is evident from Zero-shot (e.g., Phi4-14B achieving 0.75) to Prompt B (up to 0.73 for some models), yet the associated variance also increases markedly (see Figure~\ref{fig:prompting_strategies}). This suggests that while Prompt B can boost performance, it does so at the cost of reliability. Conversely, in the Albucasis dataset, despite an overall high variability across prompting configurations, Prompt A emerges as the more balanced strategy. The data in Figure~\ref{fig:combined} indicate that competitive results are attained by combinations such as Phi4-14B in both Zero-shot and Prompt A modes, COLaF’s baseline performance, as well as Qwen2.5-14B and Gemma2 when used with Prompt B. These observations underscore that the optimal prompting strategy is highly contingent on dataset-specific properties.

\begin{table*}[!ht]
\centering
\resizebox{0.7\textwidth}{!}{
    \begin{tabular}{l c ccc ccc}
    \toprule
    \multirow{2}{*}{\textbf{POS Class}} & \multirow{2}{*}{\textbf{Accuracy}} & \multicolumn{2}{c}{\textbf{Precision}} & \multicolumn{2}{c}{\textbf{Recall}} & \multicolumn{2}{c}{\textbf{F1-score}} \\
    \cmidrule(lr){3-4} \cmidrule(lr){5-6} \cmidrule(lr){7-8}
     &  & NAF6195 & Albucasis & NAF6195 & Albucasis & NAF6195 & Albucasis \\
    \midrule
    ADJ   & 0.60 & 0.60 & 0.49 & 0.58 & 0.53 & 0.59 & 0.50 \\
    ADP   & 0.79 & 0.86 & 0.95 & 0.79 & 0.74 & \cellcolor[HTML]{C6EFCE}\textbf{0.81} & 0.83 \\
    ADV   & 0.51 & 0.53 & 0.51 & 0.38 & 0.53 & 0.42 & 0.51 \\
    AUX   & 0.58 & 0.41 & 0.49 & \cellcolor[HTML]{C6EFCE}\textbf{0.91} & 0.71 & 0.39 & 0.55 \\
    CCONJ & 0.77 & \cellcolor[HTML]{C6EFCE}\textbf{0.94} & 0.95 & 0.62 & 0.79 & 0.74 & \cellcolor[HTML]{C6EFCE}\textbf{0.85} \\
    DET   & 0.78 & 0.59 & 0.72 & 0.71 & 0.79 & 0.63 & 0.75 \\
    INTJ  & 0.11 & \cellcolor[HTML]{FFC7CE}\textbf{0.00} & 0.11 & \cellcolor[HTML]{FFC7CE}\textbf{0.06} & 0.27 & \cellcolor[HTML]{FFC7CE}\textbf{0.00} & 0.13 \\
    NOUN  & 0.83 & 0.77 & 0.84 & 0.76 & \cellcolor[HTML]{C6EFCE}\textbf{0.80} & 0.76 & 0.81 \\
    NUM   & 0.69 & 0.47 & 0.61 & 0.39 & 0.75 & 0.39 & 0.65 \\
    PRON  & 0.47 & 0.57 & 0.71 & 0.40 & 0.46 & 0.46 & 0.53 \\
    PROPN & 0.48 & 0.42 & 0.12 & 0.45 & 0.59 & 0.42 & \cellcolor[HTML]{FFC7CE}\textbf{0.10} \\
    PUNCT & \cellcolor[HTML]{C6EFCE}\textbf{0.99} & 0.72 & \cellcolor[HTML]{C6EFCE}\textbf{0.99} & 0.59 & 0.58 & 0.56 & 0.70 \\
    SCONJ & 0.64 & 0.37 & 0.60 & 0.68 & 0.61 & 0.43 & 0.57 \\
    VERB  & 0.65 & 0.81 & 0.75 & 0.68 & 0.57 & 0.71 & 0.64 \\
    X     & \cellcolor[HTML]{FFC7CE}\textbf{0.03} & -- & \cellcolor[HTML]{FFC7CE}\textbf{0.01} & -- & \cellcolor[HTML]{FFC7CE}\textbf{0.02} & -- & \cellcolor[HTML]{FFC7CE}\textbf{0.01} \\
    \bottomrule
\end{tabular}

    }
\caption{Aggregated performance on UD POS tagging classes across datasets, models, and prompting strategies. The highest scores are highlighted in \colorbox[HTML]{C6EFCE}{\textbf{green}}, while lowest scores are highlighted in \colorbox[HTML]{FFC7CE}{\textbf{red}}.}
\label{tab:pos_performance}
\end{table*}

\subsection{POS Class-Level Insights}
A more granular analysis is provided by the performance metrics on the POS-Tagging class level (Table~\ref{tab:pos_performance}). High-frequency tags such as NOUN and VERB are consistently identified with accuracies of 0.83 and 0.65, respectively, and benefit from robust micro-averaged scores. In contrast, low-frequency tags such as INTJ yield extremely low accuracies (0.11 on NAF6195) and F1-scores that frequently approach zero, indicating a systemic difficulty in recognizing these classes. Moreover, classes like AUX and PROPN exhibit considerable discrepancies between macro- and micro-averaged metrics, hinting at a performance imbalance where errors in infrequent classes are overshadowed by successes in common ones.

\subsection{Model Size and Sensitivity Effects}
Our study also examines the effect of model scale on tagging performance. Models with larger parameter counts, such as Phi4-14B (14 billion parameters), generally outperform smaller counterparts like Aya-8B (8 billion parameters) across several metrics. Nonetheless, this relationship is moderated by the choice of prompting strategy as well as the supported languages (as illustrated in Table \ref{tab:supported_languages}). Sensitivity analyses (Figures~\ref{fig:model_sensitivity} and~\ref{fig:model_sensitivity_f1}) reveal that models including Mistral-7B, Mistral-Nemo-12B, and Aya-8B display heightened responsiveness to the selected prompting configuration, leading to pronounced fluctuations in accuracy and F1-score.

\subsection{Interplay Between Model Architecture and Data Characteristics}
A deeper dive into the inter-model performance reveals that models pre-trained on related high-resource languages (e.g., French, Spanish) exhibit improved robustness when applied to Old Occitan. This is particularly evident in the performance of Phi4-14B and COLaF, which not only deliver competitive results in the Zero-shot setup but also maintain stability when prompted. The variability seen in models like Mistral-7B, especially with Prompt B in the NAF6195 dataset, suggests that the underlying architecture and pre-training corpus substantially influence model behavior in low-resource settings. Trends depicted in Appendix Figures~\ref{fig:acc_vs_prompt} and~\ref{fig:f1_vs_prompt} further corroborate that both model and dataset characteristics jointly determine performance.

\section{Error Analysis}
\label{sec:erroranalysis}

A rigorous error analysis was conducted to elucidate the underlying causes of misclassifications and to identify trends that could inform future improvements. 

\subsection{POS Class-Specific Error Dynamics}
Analysis of Table~\ref{tab:pos_performance} reveals a marked disparity in performance across different POS classes. High-frequency classes such as NOUN and ADP generally yield high precision and recall; however, classes like INTJ and AUX exhibit critical shortcomings. For instance, the INTJ category in NAF6195 shows an accuracy of merely 0.11, with Precision and Recall values that fail to reach operational thresholds. Such underperformance is indicative of the insufficient representation of these classes during training, compounded by their inherent linguistic ambiguity. Additionally, classes like PROPN display a stark contrast between the two datasets—where Albucasis records a precision as low as 0.12 compared to a higher value in NAF6195—suggesting that contextual or corpus-specific factors play a predominant role in POS class classification.

\subsection{Dataset-Specific Error Patterns}
The divergence in error profiles between NAF6195 and Albucasis is noteworthy. The NAF6195 dataset’s challenging orthographic variations lead to lower overall scores, particularly affecting tags that rely on morphological subtleties (e.g., ADJ, ADV). The higher proportion of unknown vocabulary in NAF6195 exacerbates misclassification rates, as evidenced by lower Recall and F1-scores across multiple classes. Conversely, while Albucasis exhibits a generally higher baseline performance, its variability remains high; this is particularly evident when contrasting the more stable outcomes from Prompt A with the erratic performance of Prompt B. Such dataset-specific discrepancies might indicate the necessity for tailored pre-processing and normalization strategies, especially for texts with non-standard orthography.

\subsection{Cross-lingual transfer and input modifications} A striking outcome of our analysis is that the best-performing model, Phi4, achieves superior POS tagging accuracy despite modifying the input text more frequently and occasionally omitting certain words. In contrast, Mistral—although it tends to preserve the input text more faithfully—consistently exhibits lower accuracy. Phi4 has been trained on multilingual corpora, and our results (cf. Table \ref{tab:insight} ) suggest that it leverages its exposure to Romance languages (including modern Occitan) more effectively, indicating a case of Cross-lingual Transfer Learning (CLTL). Intuitively, one might expect that higher rates of textual modification or omission would yield poor performance; however, the behavior of Phi4 indicates that strategic alterations, informed by multilingual training data, can result in accurate classifications. An illustrative example is the term \emph{ancian} (english: elderly), which Mistral retains in its original form but misclassifies, whereas Phi4 transforms it into \emph{ancià} (from Catalan) and correctly classifies it. This underscores the potential of CLTL, together with prompt engineering strategies that minimize omissions, such as Zero-shot and Prompt A.

\subsection{Impact of Prompting Variability on Errors}
The choice of prompting strategy considerably affects error propagation. In the NAF6195 dataset, while Prompt B occasionally produces higher median accuracies, it also results in a larger spread of errors, as seen in the increased variance of accuracy (Figure \ref{fig:acc_vs_prompt}). This instability is less pronounced in Zero-shot and Prompt A configurations, which consistently produce more reliable outputs. In models with higher sensitivity—specifically Mistral-7B, Mistral-Nemo-12B, and Aya-8B—errors are further magnified when suboptimal prompting is employed. The analysis thus suggests that a careful balance must be struck between leveraging the potential gains of a targeted prompt and maintaining overall model robustness.

\begin{table*}[ht!]
    \centering
    \resizebox{0.9\textwidth}{!}{
    \begin{tabular}{llcccccc}
\hline
\textbf{Dataset} & \textbf{Model} & \textbf{Prompt} & \textbf{Average Levenshtein} & \textbf{Proportion Changed} & \textbf{Proportion Missing} & \textbf{Average Accuracy} \\
\hline
\multirow{6}{*}{NAF6195} 
 & \multirow{3}{*}{Mistral-7B} & Zero-shot & 0,97 & 0,06 & 0,02 & 0,65 \\
 &                         & Prompt A  & 0,97 & 0,05 & 0,02 & 0,63 \\
 &                         & Prompt B  & 0,96 & 0,07 & 0,03 & 0,59 \\
\cline{2-7}
 & \multirow{3}{*}{Phi4-14B}   & Zero-shot & 0,91 & 0,15 & 0,07 & 0,75 \\
 &                         & Prompt A  & 0,84 & 0,23 & 0,13 & 0,75 \\
 &                         & Prompt B  & 0,87 & 0,20 & 0,11 & 0,73 \\
\hline
\multirow{6}{*}{Albucasis}
 & \multirow{3}{*}{Mistral-7B} & Zero-shot & 0,94 & 0,10 & 0,05 & 0,70 \\
 &                          & Prompt A  & 0,94 & 0,11 & 0,05 & 0,71 \\
 &                          & Prompt B  & 0,91 & 0,13 & 0,08 & 0,72 \\
\cline{2-7}
 & \multirow{3}{*}{Phi4-14B}    & Zero-shot & 0,90 & 0,15 & 0,08 & 0,84 \\
 &                          & Prompt A  & 0,87 & 0,19 & 0,11 & 0,82 \\
 &                          & Prompt B  & 0,86 & 0,20 & 0,12 & 0,80 \\
\hline
\end{tabular}
    }
    \caption{Comparison of Phi4-14B and Mistral-7B in terms of ratio of changes of original input text, ratio of omissions and average accuracy, across the NAF6195 and \textit{Albucasis} datasets.}
    \label{tab:insight}
\end{table*}

\subsection{Error Propagation Across Model Architectures}
Our sensitivity analysis, as depicted in Figures~\ref{fig:model_sensitivity} and~\ref{fig:model_sensitivity_f1}, indicates that the propagation of errors is not uniformly distributed across model architectures. Larger models such as Phi4-14B tend to contain errors within lower-frequency POS classes, whereas smaller or more sensitive models show a broader dispersion of misclassifications. The inherent variability in performance, particularly under Prompt B conditions, suggests that model architecture and pre-training corpus composition are critical determinants of error propagation in low-resource language processing.

\section{Practical Recommendations}
\label{sec:recommendations}

Drawing on the detailed results and error analyses, we propose several recommendations to optimize POS tagging in Old Occitan. Our suggestions address model selection, pre-processing strategies, and the tuning of prompting configurations.

\subsection{Pre-processing and CLTL}
To address the challenges posed by non-standard orthography and high rates of unknown vocabulary, solutions such as integrating pre-processing pipelines might be considered. Techniques such as orthographic normalization, vocabulary expansion using external resources like the DOM (Dictionnaire de l'Occitan Mediéval), and context-aware tokenization are recommended. Further, we observe that models that are exposed to languages of the same family tend to exhibit higher robustness toward spelling and prompting variations. These steps might reduce error rates in classes that require subtle morphological distinctions and improve overall tagging performance.

\subsection{Optimizing Prompting Strategies}
The data clearly indicate that the choice of prompting strategy influences model outcomes substantially. For datasets with high orthographic variability, such as NAF6195, while Prompt B can offer higher median accuracy, its increased variance necessitates cautious deployment. In contrast, Prompt A has demonstrated a better balance between performance and stability in Albucasis. Practitioners are advised to experiment with multiple prompting configurations during development and to select the one that offers the best trade-off between accuracy and consistency. Furthermore, automated prompt tuning and cross-validation across multiple runs can help in identifying the most robust configuration for a given dataset.

\subsection{Model Selection and Configuration}
For practitioners aiming to deploy robust POS tagging systems, our findings recommend prioritizing models that demonstrate consistent performance across both Zero-shot and prompted configurations. Models like Phi4-14B and COLaF exhibit superior performance and stability, making them prime candidates for further refinement. Given that larger models tend to perform better but may incur higher computational costs, the choice should balance resource availability with performance needs. Sensitivity analyses further suggest avoiding overly sensitive models, such as Mistral-7B and Aya-8B, unless ensemble methods or targeted fine-tuning strategies are employed to mitigate their variability.

\section{Conclusion}
\label{conclusion}

This study provides the first systematic evaluation of LLMs for POS tagging in Old Occitan, a highly non-standardized and low-resource historical language. Our findings reveal that while larger models demonstrate some ability to generalize, all tested LLMs struggle with morphological and syntactic inconsistencies due to the lack of training data in similar linguistic contexts. Prompting strategies such as few-shot learning show potential for improving tagging accuracy, yet challenges remain in fine-tuning models for historical text understanding. Furthermore, our error analysis highlights specific areas where LLMs perform poorly, such as handling orthographic variation and a low degree of cross-lingual transfer learning. The insights gained from this work pave the way for further research in historical NLP, emphasizing the need for better-prepared training datasets and refined evaluation methodologies tailored to non-standardized languages. In future work, we plan to extend our analysis to other low-resource languages, including Old French and Medieval Latin, and evaluate the effect of fine-tuning and choice of decoding strategies over the POS tagging quality.

\section*{Limitations}
\label{limitations}

While this study offers valuable insights into the application of modern natural language processing techniques to historical, low-resource languages, several limitations must be acknowledged.

\noindent
Firstly, the analysis is based on a dataset comprised solely of archival Old Occitan texts. Despite considerable efforts to expand the corpus of Old Occitan material \citep{garces-arias-etal-2025-decoding}, the inherent scarcity of such sources inevitably constrains the generalisability of our findings.

\noindent
Secondly, our evaluation was restricted to eight open‐source models. Consequently, the performance and potential of additional architectures—and notably, proprietary models—remain to be assessed.

\noindent
Thirdly, our choice of open-source models was additionally limited due to the hardware requirements. Larger models like Llama 3.3 could therefore not be investigated.  

\noindent
Fourthly, although three prompting strategies of progressively increasing complexity were explored, alternative prompting designs merit further investigation. In particular, the impacts of varying tokenization procedures and the potential benefits of fine-tuning with dedicated Old Occitan corpora are avenues for future research.

\noindent
Finally, the influence of decoding strategies on the quality of part-of-speech tagging predictions was not fully explored, representing an additional dimension for subsequent studies.

\section*{Ethics Statement}

We affirm that our research adheres to the \href{https://www.aclweb.org/portal/content/acl-code-ethics}{ACL Ethics Policy}. This work involves the use of publicly available datasets and does not involve human
subjects or any personally identifiable information.
We declare that we have no conflicts of interest that
could potentially influence the outcomes, interpretations, or conclusions of this research. All funding
sources supporting this study are acknowledged in
the acknowledgments section. We have made our
best effort to document our methodology, experiments, and results accurately and are committed to
sharing our code, data, and other relevant resources
to foster reproducibility and further advancements
in research.

\section*{Acknowledgments}

We would like to express our sincere gratitude to Viola Baltzer and Verena Harrer for their valuable assistance in preparing and annotating our datasets. Matthias Aßenmacher gratefully acknowledges financial support provided by the Deutsche Forschungsgemeinschaft (DFG, German Research Foundation) through the BERD@NFDI project, grant number 460037581. Additionally, we thank the Leibniz-Rechenzentrum der Bayerischen Akademie der Wissenschaften (LRZ) for providing computational resources essential for this research.

\bibliography{custom}

\clearpage
\onecolumn

\appendix

\section{Appendix}
\label{sec:appendix}

\subsection{Metrics}
\label{a:metrics}

\paragraph{Accuracy}
Accuracy measures the proportion of correctly predicted POS tags over the total number of tags:
\begin{equation}
    \text{Accuracy} = \frac{TP + TN}{TP + TN + FP + FN}
\end{equation}
where \(TP\), \(TN\), \(FP\), and \(FN\) represent true positives, true negatives, false positives, and false negatives, respectively.

\paragraph{Precision}
Precision evaluates the proportion of correctly predicted POS tags among all predicted instances of a given tag:
\begin{equation}
    \text{Precision} = \frac{TP}{TP + FP}
\end{equation}

\paragraph{Recall}
Recall measures the proportion of correctly predicted POS tags out of all actual instances of that tag:
\begin{equation}
    \text{Recall} = \frac{TP}{TP + FN}
\end{equation}

\paragraph{F1-score}
The F1-score provides a balance between precision and recall and is defined as:
\begin{equation}
    \text{F1-score} = 2 \times \frac{\text{Precision} \times \text{Recall}}{\text{Precision} + \text{Recall}}
\end{equation}

\paragraph{Averaging in a Multiclass Setting}
Given that POS tagging is a multiclass task, the evaluation metrics are computed using different averaging strategies:
\begin{itemize}
    \item \textbf{Micro Averaging:} This method aggregates the contributions of all classes by summing the individual true positives, false positives, and false negatives across all classes. The metrics are then computed from these global counts. As a result, micro averaging is particularly sensitive to the performance on frequent classes.
    \item \textbf{Macro Averaging:} In this approach, the metric is computed independently for each class, and the final score is obtained by taking the arithmetic mean of these per-class metrics. This gives equal weight to each class, thus emphasizing performance on both common and rare classes.
    \item \textbf{Weighted Averaging:} Here, each class's metric is weighted by its support (i.e., the number of true instances). The overall metric is computed as a weighted average of the individual class scores, thereby reflecting the class distribution in the dataset.
\end{itemize}
\paragraph{RCPTP: Ratio of Correctly POS-Tagged Phrases} This metric measures the proportion of phrases without POS Tagging errors:
\begin{equation}
    \text{RCPTP} = \frac{\text{Number of correct phrases}}{\text{Total number of phrases}}
\end{equation}
This metric provides insights into how well LLMs refine and improve initial POS tagging predictions.
\noindent
Note that the term \textit{sentence} or \textit{phrase} is highly ambiguous; we find many different definitions ranging from purely pragmatical or semantical approaches to graphical or intonational definitions \cite{mieszkowski2019crises}. For the purpose of this paper, we employed a syntactical definition based on punctuation: all words between two periods are seen as belonging to one phrase.

\clearpage

\subsection{Prompting Strategies}

\begin{table*}[ht!]
    \centering
    \begin{tabular}{|l|p{12cm}|}
        \hline
        \textbf{Prompting Strategy} & \textbf{Prompt} \\
        \hline
        \textbf{Zero-shot} & Analyze the provided text and assign to each word Universal Dependencies Part-of-Speech tags: ``ADJ", ``ADP", ``ADV", ``AUX", ``CCONJ", ``DET", ``INTJ", ``NOUN", ``NUM", ``PRON", ``PROPN", ``PUNCT", ``SCONJ", ``VERB", ``X". Return the results as a JSON array of objects, each containing only the 'word' and 'upos' keys. The output must be only the JSON array without any additional text, explanations, or formatting. \\
        \hline
        \textbf{Prompt A} & \textit{You are a Medieval Occitan language expert. Analyze the provided text and assign to each word Universal Dependencies Part-of-Speech tags: ``ADJ", ``ADP", ``ADV", ``AUX", ``CCONJ", ``DET", ``INTJ", ``NOUN", ``NUM", ``PRON", ``PROPN", ``PUNCT", ``SCONJ", ``VERB", ``X". Do not add or remove punctuation or tokens. Ensure to process token by token. Ensure that the order of words in the text is kept for the output. Return the results as a JSON array of objects, each containing only the 'word' and 'upos' keys. The output must be only the JSON array without any additional text, explanations, or formatting. Ensure that the JSON array is properly closed.} \\
        \hline
        \textbf{Prompt B} & \textit{You are a medieval Occitan language expert specializing in linguistic analysis. This language is related to Catalan and Latin. In this text there is a high variety of spelling variations having the same meaning} \\ 
        \textbf{} & \textit{This is an example for spelling variation:} \\
        \textbf{} & \textit{homps, ome, om, omen, omne, hom, home.} \\
        \textbf{} & \textit{Another example is:} \\
        \textbf{} & \textit{acayson, achaison, acheison, acheson, aqueison, caiso, caison, cason, cayson, chaizo, queison or gaug, gauc, gautz, jau, jauvi.} \\
        \textbf{} & \textit{Your task is to analyze the given text and assign Universal Dependencies Part-of-Speech (UD POS) tags to each word.} \\
        \textbf{} & \textit{Return the results as a JSON array of objects, each containing only the 'word' and 'upos' keys.} \\
        \textbf{} & \textit{Ensure that the JSON array is properly formatted and closed.} \\
        \textbf{} & \textit{The output must be only the JSON array without any additional text, explanations, or formatting} \\
        \hline
    \end{tabular}
    \caption{Comparison of different prompting strategies for UD POS tagging}
    \label{tab:prompting_strategies}
\end{table*}

\FloatBarrier  

\clearpage

\subsection{Dataset POS Tagging performance}
\label{postag_performance_dataset}

\begin{table*}[ht!]
    \centering
    \resizebox{0.7\textwidth}{!}{
    \begin{tabular}{l c ccc ccc ccc}
    \toprule
    \multirow{2}{*}{\textbf{Model}} & \multirow{2}{*}{\textbf{Accuracy}} & \multicolumn{3}{c}{\textbf{Precision}} & \multicolumn{3}{c}{\textbf{Recall}} & \multicolumn{3}{c}{\textbf{F1-score}} \\
    \cmidrule(lr){3-5} \cmidrule(lr){6-8} \cmidrule(lr){9-11}
    & & micro & macro & wavg & micro & macro & wavg & micro & macro & wavg \\
    \midrule
    COLaF & 0.66 & 0.66 & 0.60 & 0.67 & 0.66 & 0.61 & 0.66 & 0.66 & 0.58 & 0.65 \\
    \midrule
    Phi4-14B (Zero-shot) & \cellcolor[HTML]{C6EFCE}\textbf{0.75} & \cellcolor[HTML]{C6EFCE}\textbf{0.75} & \cellcolor[HTML]{C6EFCE}\textbf{0.65} & \cellcolor[HTML]{C6EFCE}\textbf{0.77} & \cellcolor[HTML]{C6EFCE}\textbf{0.75} & 0.68 & \cellcolor[HTML]{C6EFCE}\textbf{0.75} & \cellcolor[HTML]{C6EFCE}\textbf{0.75} & \cellcolor[HTML]{C6EFCE}\textbf{0.66} & \cellcolor[HTML]{C6EFCE}\textbf{0.75} \\
    Phi4-14B (Prompt A) & \cellcolor[HTML]{C6EFCE}\textbf{0.75} & \cellcolor[HTML]{C6EFCE}\textbf{0.75} & 0.64 & 0.76 & \cellcolor[HTML]{C6EFCE}\textbf{0.75} & 0.67 & \cellcolor[HTML]{C6EFCE}\textbf{0.75} & \cellcolor[HTML]{C6EFCE}\textbf{0.75} & 0.64 & 0.74 \\
    Phi4-14B (Prompt B) & 0.73 & 0.73 & 0.63 & 0.75 & 0.73 & 0.62 & 0.61 & 0.73 & 0.61 & 0.73 \\
    \midrule
    Mistral-7B (Zero-shot) & 0.65 & 0.65 & 0.55 & 0.67 & 0.65 & 0.58 & 0.65 & 0.65 & 0.56 & 0.65 \\
    Mistral-7B (Prompt A) & 0.63 & 0.63 & 0.55 & 0.67 & 0.63 & 0.56 & 0.63 & 0.63 & 0.54 & 0.64 \\
    Mistral-7B (Prompt B) & 0.59 & 0.59 & \cellcolor[HTML]{FFC7CE}\textbf{0.48} & \cellcolor[HTML]{FFC7CE}\textbf{0.62} & 0.59 & 0.47 & 0.59 & 0.59 & \cellcolor[HTML]{FFC7CE}\textbf{0.41} & 0.59 \\
    \midrule
    Mistral-Nemo-12B (Zero-shot) & 0.66 & 0.66 & 0.53 & 0.71 & 0.66 & 0.59 & 0.66 & 0.66 & 0.51 & 0.67 \\
    Mistral-Nemo-12B (Prompt A) & 0.69 & 0.69 & 0.60 & 0.73 & 0.69 & \cellcolor[HTML]{C6EFCE}\textbf{0.69} & 0.68 & 0.69 & 0.58 & 0.69 \\
    Mistral-Nemo-12B (Prompt B) & 0.68 & 0.68 & 0.54 & 0.71 & 0.68 & 0.60 & 0.68 & 0.68 & 0.51 & 0.68 \\
    \midrule
    Gemma2-9B (Zero-shot) & 0.65 & 0.65 & 0.50 & 0.68 & 0.65 & 0.55 & 0.65 & 0.65 & 0.48 & 0.65 \\
    Gemma2-9B (Prompt A) & 0.68 & 0.68 & 0.55 & 0.70 & 0.68 & 0.58 & 0.67 & 0.68 & 0.55 & 0.68 \\
    Gemma2-9B (Prompt B) & 0.70 & 0.70 & \cellcolor[HTML]{C6EFCE}\textbf{0.65} & 0.72 & 0.70 & 0.60 & 0.70 & 0.70 & 0.60 & 0.69 \\
    \midrule
    Mixtral-8x7B (Zero-shot) & 0.65 & 0.65 & 0.60 & 0.69 & 0.65 & 0.56 & 0.65 & 0.65 & 0.56 & 0.66 \\
    Mixtral-8x7B (Prompt A) & 0.67 & 0.67 & 0.56 & 0.70 & 0.67 & 0.57 & 0.67 & 0.67 & 0.55 & 0.68 \\
    Mixtral-8x7B (Prompt B) & 0.67 & 0.67 & 0.59 & 0.70 & 0.67 & 0.57 & 0.67 & 0.67 & 0.57 & 0.68 \\
    \midrule
    Aya-8B (Zero-shot) & 0.60 & 0.60 & 0.50 & 0.67 & 0.60 & \cellcolor[HTML]{FFC7CE}\textbf{0.46} & 0.60 & 0.60 & 0.44 & 0.62 \\
    Aya-8B (Prompt A) & 0.61 & 0.61 & 0.53 & 0.66 & 0.61 & 0.56 & 0.61 & 0.61 & 0.52 & 0.62 \\
    Aya-8B (Prompt B) & \cellcolor[HTML]{FFC7CE}\textbf{0.57} & \cellcolor[HTML]{FFC7CE}\textbf{0.57} & 0.52 & 0.65 & \cellcolor[HTML]{FFC7CE}\textbf{0.57} & 0.52 & \cellcolor[HTML]{FFC7CE}\textbf{0.57} & \cellcolor[HTML]{FFC7CE}\textbf{0.57} & 0.49 & \cellcolor[HTML]{FFC7CE}\textbf{0.58} \\
    \midrule
    Qwen2.5-14B (Zero-shot) & 0.66 & 0.66 & 0.60 & 0.72 & 0.66 & 0.59 & 0.66 & 0.66 & 0.56 & 0.67 \\
    Qwen2.5-14B (Prompt A) & 0.70 & 0.70 & 0.63 & 0.75 & 0.70 & 0.64 & 0.70 & 0.70 & 0.61 & 0.71\\
    Qwen2.5-14B (Prompt B) & 0.72 & 0.72 & \cellcolor[HTML]{C6EFCE}\textbf{0.65} & 0.75 & 0.72 & 0.61 & 0.72 & 0.72 & 0.61 & 0.71 \\
    \bottomrule
\end{tabular}
    }
    \caption{Average scores across all models for the NAF6195 dataset. The highest scores are highlighted in \colorbox[HTML]{C6EFCE}{\textbf{green}}, while lowest scores are highlighted in \colorbox[HTML]{FFC7CE}{\textbf{red}}.}
    \label{tab:pos_tagging_performance_naf6195}
\end{table*}

\begin{table*}[ht!]
    \centering
    \resizebox{0.7\textwidth}{!}{
    \begin{tabular}{l c ccc ccc ccc}
    \toprule
    \multirow{2}{*}{\textbf{Model}} & \multirow{2}{*}{\textbf{Accuracy}} & \multicolumn{3}{c}{\textbf{Precision}} & \multicolumn{3}{c}{\textbf{Recall}} & \multicolumn{3}{c}{\textbf{F1-score}} \\
    \cmidrule(lr){3-5} \cmidrule(lr){6-8} \cmidrule(lr){9-11}
    & & micro & macro & wavg & micro & macro & wavg & micro & macro & wavg \\
    \midrule
    COLaF & 0.80 & 0.80 & 0.61 & 0.81 & 0.80 & 0.65 & 0.80 & 0.80 & 0.61 & 0.80 \\
    \midrule
    Phi4-14B (Zero-shot) & \cellcolor[HTML]{C6EFCE}\textbf{0.84} & \cellcolor[HTML]{C6EFCE}\textbf{0.84} & 0.67 & \cellcolor[HTML]{C6EFCE}\textbf{0.87} & \cellcolor[HTML]{C6EFCE}\textbf{0.84} & \cellcolor[HTML]{C6EFCE}\textbf{0.77} & \cellcolor[HTML]{C6EFCE}\textbf{0.84} & 0.84 & 0.69 & \cellcolor[HTML]{C6EFCE}\textbf{0.85} \\
    Phi4-14B (Prompt A) & 0.82 & 0.82 & 0.67 & 0.85 & 0.82 & 0.74 & 0.82 & 0.82 & 0.67 & 0.83 \\
    Phi4-14B (Prompt B) & 0.80 & 0.80 & 0.65 & 0.82 & 0.80 & 0.73 & 0.80 & 0.80 & 0.66 & 0.80 \\
    \midrule
    Mistral-7B (Zero-shot) & 0.70 & 0.70 & 0.57 & 0.75 & 0.70 & 0.63 & 0.70 & 0.70 & 0.55 & 0.70 \\
    Mistral-7B (Prompt A) & 0.71 & 0.71 & 0.55 & 0.76 & 0.71 & 0.64 & 0.71 & 0.71 & 0.54 & 0.72 \\
    Mistral-7B (Prompt B) & 0.72 & 0.72 & 0.63 & 0.77 & 0.72 & 0.58 & 0.72 & 0.72 & 0.56 & 0.73 \\
    \midrule
    Mistral-Nemo-12B (Zero-shot) & 0.71 & 0.71 & 0.57 & 0.75 & 0.71 & 0.68 & 0.71 & 0.71 & 0.58 & 0.72 \\
    Mistral-Nemo-12B (Prompt A) & 0.76 & 0.76 & 0.62 & 0.82 & 0.76 & 0.66 & 0.76 & 0.76 & 0.57 & 0.76 \\
    Mistral-Nemo-12B (Prompt B) & \cellcolor[HTML]{FFC7CE}\textbf{0.67} & \cellcolor[HTML]{FFC7CE}\textbf{0.67} & 0.54 & \cellcolor[HTML]{FFC7CE}\textbf{0.74} & \cellcolor[HTML]{FFC7CE}\textbf{0.67} & 0.65 & \cellcolor[HTML]{FFC7CE}\textbf{0.67} & \cellcolor[HTML]{FFC7CE}\textbf{0.66} & 0.56 & \cellcolor[HTML]{FFC7CE}\textbf{0.68} \\
    \midrule
    Gemma2-9B (Zero-shot) & 0.72 & 0.72 & 0.55 & 0.75 & 0.72 & \cellcolor[HTML]{FFC7CE}\textbf{0.56} & 0.72 & 0.72 & 0.51 & 0.71 \\
    Gemma2-9B (Prompt A) & 0.75 & 0.75 & 0.57 & 0.78 & 0.75 & 0.62 & 0.75 & 0.75 & 0.55 & 0.74 \\
    Gemma2-9B (Prompt B) & 0.73 & 0.73 & 0.66 & 0.77 & 0.73 & 0.60 & 0.73 & 0.73 & 0.59 & 0.71 \\
    \midrule
    Mixtral-8x7B (Zero-shot) & 0.75 & 0.75 & 0.59 & 0.77 & 0.74 & 0.63 & 0.75 & 0.75 & 0.58 & 0.75 \\
    Mixtral-8x7B (Prompt A) & 0.78 & 0.78 & 0.60 & 0.79 & 0.78 & 0.65 & 0.78 & 0.78 & 0.60 & 0.78 \\
    Mixtral-8x7B (Prompt B) & 0.79 & 0.79 & 0.66 & 0.80 & 0.79 & 0.68 & 0.79 & 0.79 & 0.66 & 0.79 \\
    \midrule
    Aya-8B (Zero-shot) & 0.69 & 0.69 & \cellcolor[HTML]{FFC7CE}\textbf{0.49} & 0.76 & 0.69 & 0.57 & 0.69 & 0.69 & \cellcolor[HTML]{FFC7CE}\textbf{0.48} & 0.71 \\
    Aya-8B (Prompt A) & 0.71 & 0.71 & 0.57 & 0.79 & 0.71 & 0.67 & 0.71 & 0.71 & 0.56 & 0.73 \\
    Aya-8B (Prompt B) & 0.75 & 0.75 & 0.60 & 0.81 & 0.75 & 0.66 & 0.75 & 0.75 & 0.57 & 0.75 \\
    \midrule
    Qwen2.5-14B (Zero-shot) & 0.79 & 0.79 & 0.64 & 0.86 & 0.79 & 0.75 & 0.79 & \cellcolor[HTML]{C6EFCE}\textbf{0.86} & \cellcolor[HTML]{C6EFCE}\textbf{0.79} & 0.82 \\
    Qwen2.5-14B (Prompt A) & 0.77 & 0.77 & 0.60 & 0.84 & 0.77 & 0.73 & 0.77 & 0.77 & 0.59 & 0.79\\
    Qwen2.5-14B (Prompt B) & 0.79 & 0.79 & \cellcolor[HTML]{C6EFCE}\textbf{0.68} & 0.81 & 0.79 & 0.75 & 0.79 & 0.79 & 0.68 & 0.79 \\
    \bottomrule
\end{tabular}

    }
    \caption{Average scores across all models for the \textit{Albucasis} dataset. The highest scores are highlighted in \colorbox[HTML]{C6EFCE}{\textbf{green}}, while lowest scores are highlighted in \colorbox[HTML]{FFC7CE}{\textbf{red}}.}
    \label{tab:pos_tagging_performance_albucasis}
\end{table*}

\FloatBarrier  

\clearpage

\subsection{Model sensitivity}
\label{a:sensitivity}

\begin{figure*}[ht!]
    \centering
    \includegraphics[width=0.7\textwidth]{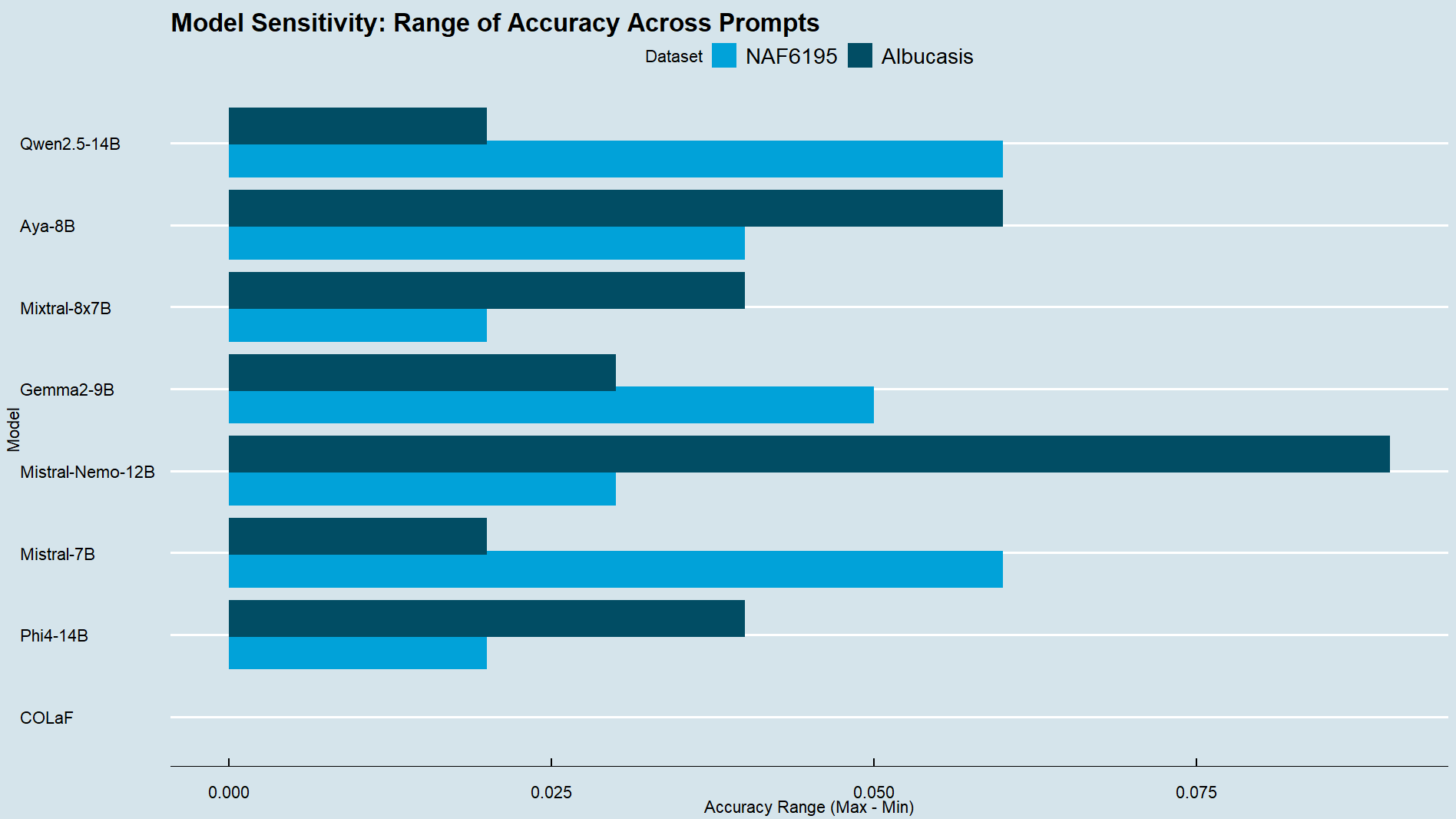}
\caption{Range of accuracy per model across prompts.}
 \label{fig:model_sensitivity} 
\end{figure*}

\begin{figure*}[ht!]
    \centering
    \includegraphics[width=0.7\textwidth]{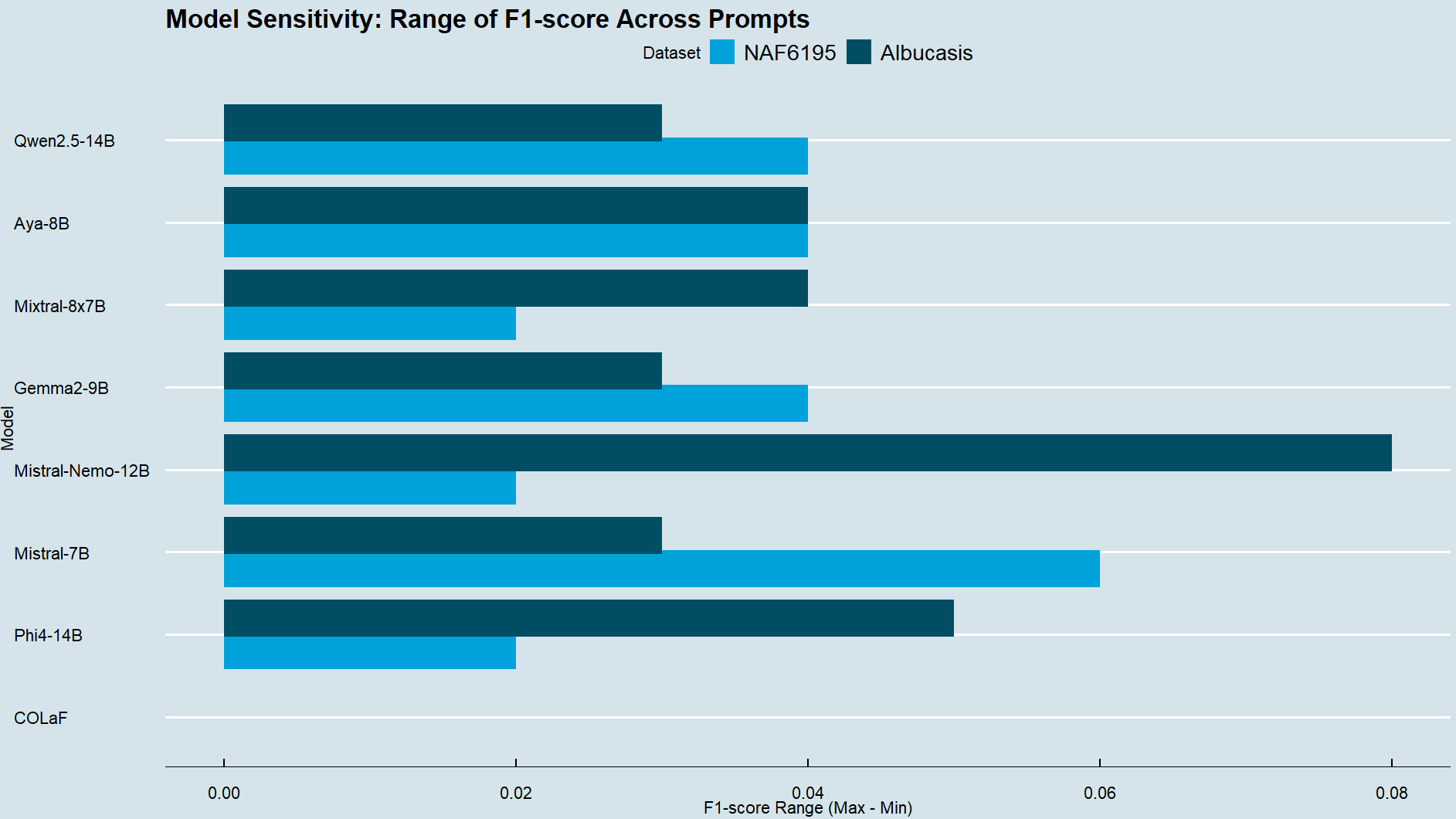}
\caption{Range of F1-score per model across prompts.}
 \label{fig:model_sensitivity_f1} 
\end{figure*}

\clearpage

\subsection{Further results}
\label{a:further_results}

\begin{figure*}[ht!]
    \centering
    \includegraphics[width=0.7\textwidth]{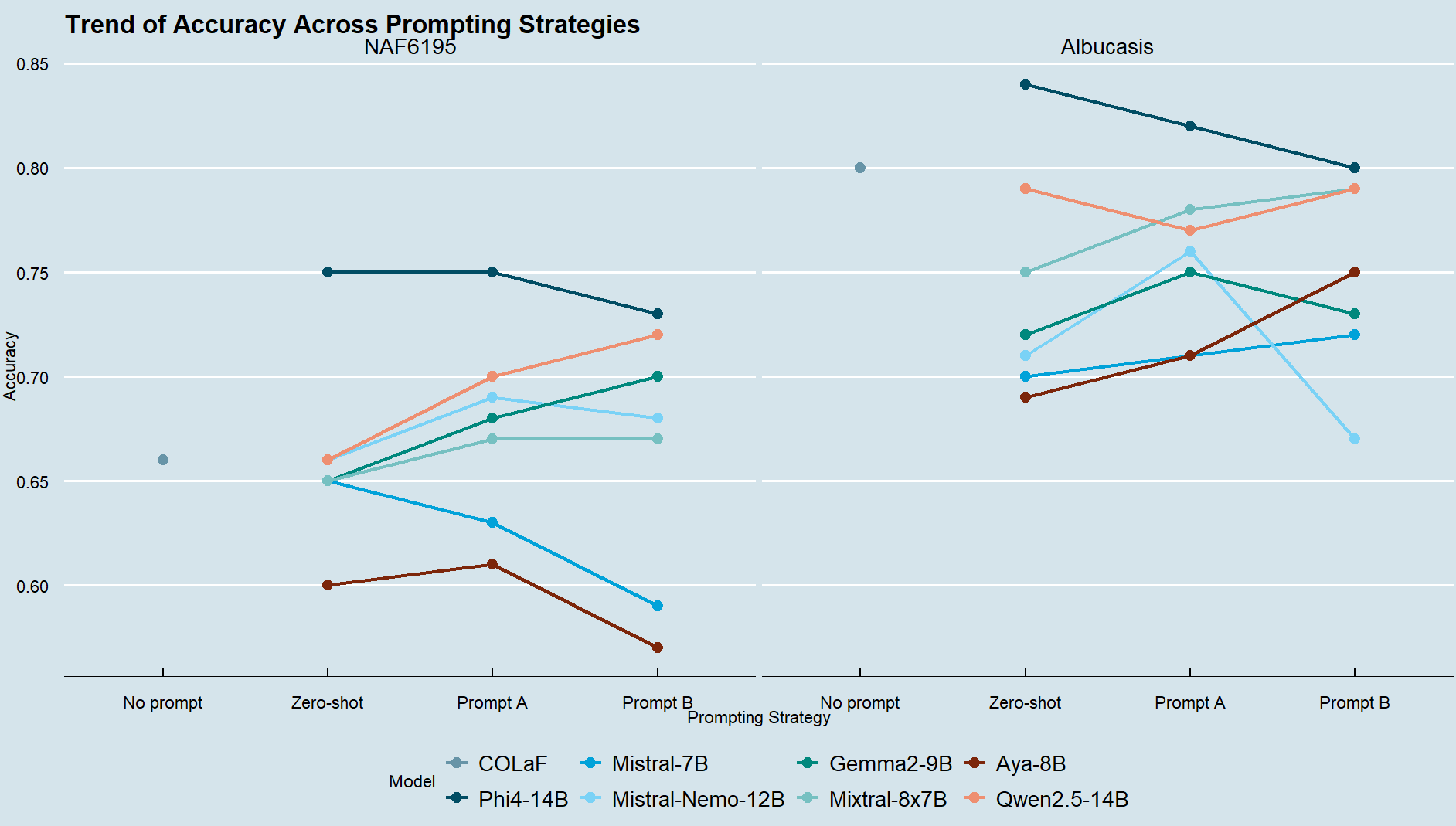}
\caption{Accuracy behavior vs. Prompting strategies.}
 \label{fig:acc_vs_prompt} 
\end{figure*}

\begin{figure*}[ht!]
    \centering
    \includegraphics[width=0.7\textwidth]{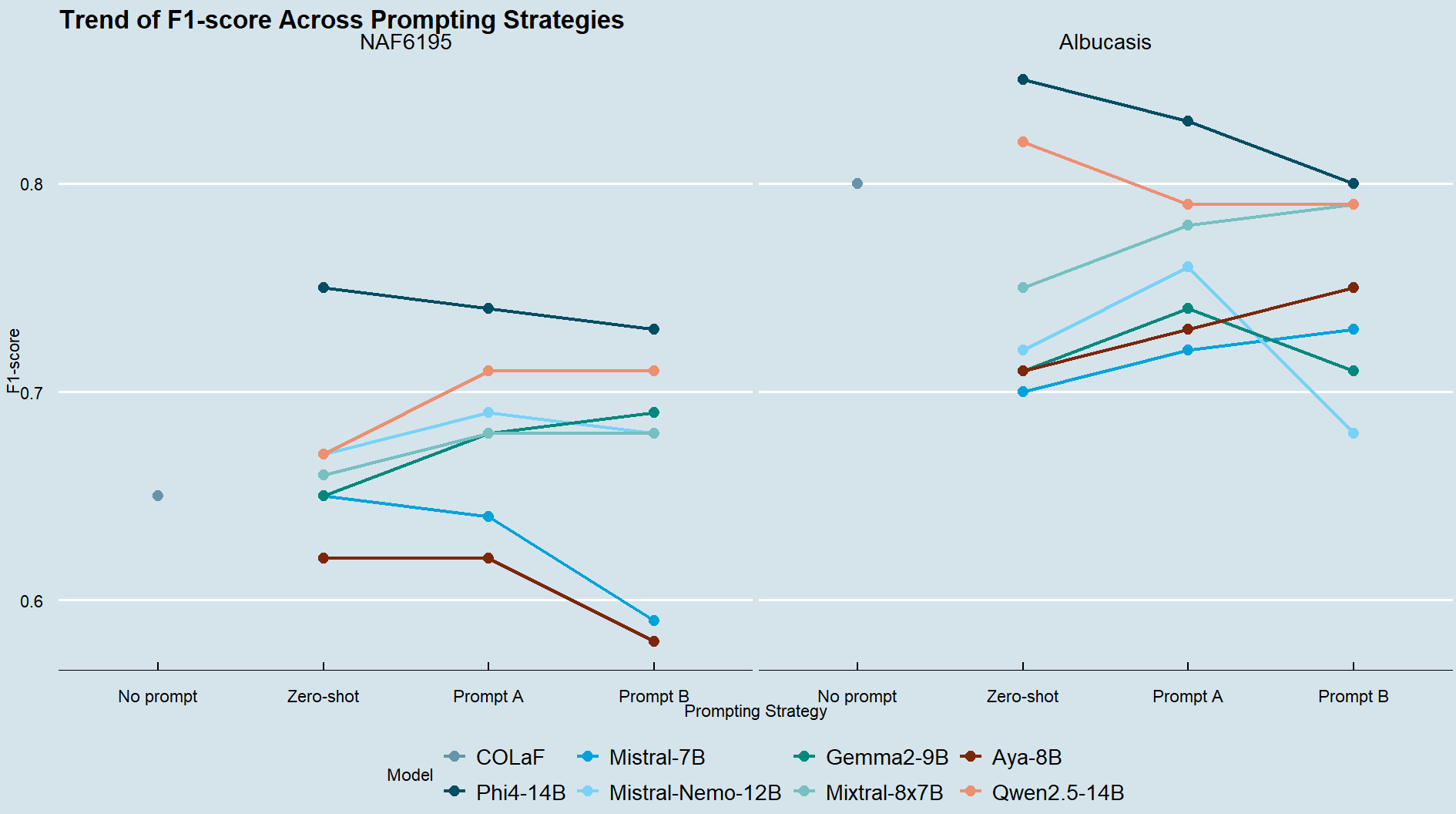}
\caption{F1-score behavior vs. Prompting strategies.}
 \label{fig:f1_vs_prompt} 
\end{figure*}

\begin{comment}
\begin{figure*}[ht!]
    \centering
    \includegraphics[width=0.7\textwidth]{relationship_metrics.png}
\caption{Relationship between Accuracy and F1-score across different prompting strategies and models.}
 \label{fig:f1_vs_accuracy} 
\end{figure*}
\end{comment}

\begin{comment}
\begin{table*}[ht!]
    \centering
    \resizebox{0.7\textwidth}{!}{
    \input{top_five_bottom_five}
    }
\caption{Top 5 and Bottom 5 Model+Prompting Strategy Combinations by Error Rate}
\label{tab:performance_extremes}

\end{table*}
\end{comment}

\end{document}